\documentclass[10pt,twocolumn,letterpaper]{article}

\usepackage{cvpr}              
\definecolor{cvprblue}{rgb}{0.21,0.49,0.74}
\usepackage[pagebackref,breaklinks,colorlinks,allcolors=cvprblue]{hyperref}
\usepackage{listings}
\def\paperID{21293} 
\def\confName{CVPR}
\def\confYear{2026}

\title{Decomposing Visual Classification: Assessing Tree-Based Reasoning in VLMs}

\author{
Sary Elmansoury\\
German Research Centre for Artificial Intelligence (DFKI)\\
Department of Data Science and its Applications\\
{\tt\small sary.elmansoury@dfki.de}
\and
Islam Mesabah\\
German Research Centre for Artificial Intelligence (DFKI)\\
Department of Data Science and its Applications\\
{\tt\small islam.mesabah@dfki.de}
\and
Gerrit Großmann\\
German Research Centre for Artificial Intelligence (DFKI)\\
Department of Data Science and its Applications\\
{\tt\small gerrit.grossmann@dfki.de}
\and
Peter Neigel\\
University of Kaiserslautern--Landau (RPTU)\\
Department of Computer Science\\
{\tt\small peter.neigel@dfki.de}
\and
Daniel Kondermann\\
Heidelberg University\\
Faculty for Mathematics and Computer Sciences\\
{\tt\small dk@quality-match.com}
\and
Sebastian Vollmer\\
German Research Centre for Artificial Intelligence (DFKI)\\
Department of Data Science and its Applications\\
{\tt\small sebastian.vollmer@dfki.de}
}

\begin{document}
\maketitle
\begin{abstract}
Vision language models (VLMs) excel at zero-shot visual classification, but their performance on fine-grained tasks and large hierarchical label spaces is understudied. This paper investigates whether structured, tree-based reasoning can enhance VLM performance. We introduce a framework that decomposes classification into interpretable decisions using decision trees and evaluates it on fine-grained (GTSRB \& MNIST) and coarse-grained (MSCOCO \& CIFAR-10) datasets. Although the model achieves 98.2\% accuracy in understanding the tree knowledge, tree-based reasoning consistently underperforms standard zero-shot prompting. We also explore enhancing the tree prompts with LLM-generated classes and image descriptions to improve alignment. The added description enhances the performance of the tree based and zero-shot methods. Our findings highlight limitations of structured reasoning in visual classification and offer insights for designing more interpretable VLM systems.
\end{abstract}

\section{Introduction}

Vision language models (VLMs) have revolutionized visual classification by leveraging powerful image-text pretraining to generalize across novel categories without task-specific supervision~\cite{radford2021learningtransferablevisualmodels, openai2024gpt4technicalreport, bai2023qwenvlversatilevisionlanguagemodel}. However, as these models are increasingly deployed in real-world applications, two critical limitations have emerged that challenge their practical utility.

First, VLMs struggle with fine-grained classification tasks involving large, hierarchically organized label spaces~\cite{zhang2024visuallygroundedlanguagemodelsbad}. While models like CLIP and GPT-4V excel at distinguishing between semantically distinct categories (e.g., "car" vs. "airplane"), they often fail when faced with subtle visual distinctions within specialized domains, such as differentiating between traffic sign variants or medical imaging subcategories~\cite{kim-ji-2024-finer, wei2024enhancingfinegrainedimageclassifications}. This limitation is particularly problematic in safety-critical applications where precise classification is essential.

Second, the black-box nature of current VLM decision-making processes limits their interpretability and debuggability. When a model misclassifies an input, practitioners have limited insight into the reasoning pathway that led to the error, making it difficult to identify systematic biases or improve model performance. This opacity is especially concerning in domains requiring accountability and transparency.

To address these challenges, recent work has proposed incorporating structured reasoning approaches, particularly hierarchical decision trees, into VLM inference pipelines~\cite{yellinek20253vlusingtreesimprove, ding2025treeattributespromptlearning}. The theoretical appeal is compelling as it suggests
 decomposing complex tasks into a series of interpretable decisions, tree-based reasoning promises to improve accuracy and explainability. Each decision node in the tree can be designed to capture specific visual or semantic distinctions. 

However, the practical effectiveness of tree-based reasoning for classification remains an open empirical question. While such methods have shown promise in tasks like image captioning, it is unclear whether the added structural complexity benefits modern VLMs in classification tasks or if it simply introduces new sources of error. Notably, the hierarchical nature of decision trees makes them prone to error propagation, where a single incorrect decision at a higher-level node can compromise the entire classification path. In addition, Hierarchical classification and interpretability, its real-world effectiveness is still under debate. Ongoing work is needed to overcome error propagation, optimize feature alignment, and validate empirical gains in diverse, safety-critical settings.

In this paper, we conduct a systematic empirical evaluation of tree-based reasoning in VLMs, comparing hierarchical decision-making against standard zero-shot image classification across multiple models, datasets, prompts and settings. Our investigation reveals a surprising finding: Contrary to theoretical expectations, tree-based reasoning consistently underperforms direct zero-shot classification, often by substantial margins. Through detailed error analysis, we identify the key factors contributing to this performance degradation, including error propagation through hierarchical structures, sensitivity to question formulation, and depth-related accuracy penalties.

Our contributions are:  
\begin{enumerate}
    \item We provide a comprehensive empirical assessment of tree-based reasoning in multiple VLMs across four classification tasks under diverse settings, which ensures the generalization of the experiments. 
    \item We identify and analyze the specific failure modes that limit the effectiveness of hierarchical reasoning approaches. 
    \item We offer design recommendations for future structured reasoning systems based on our findings. 
\end{enumerate}
These results have important implications for the development of interpretable VLM systems and highlight the drawbacks of the current usage of hierarchical reasoning methods.
\section{Related Work}

\textbf{Vision-Language Models and Zero-Shot Classification.}\\
VLMs such as CLIP~\cite{radford2021learningtransferablevisualmodels}, LLaVA~\cite{liu2023visualinstructiontuning}, Qwen-VL~\cite{bai2023qwenvlversatilevisionlanguagemodel}, and GPT-4V~\cite{openai2024gpt4technicalreport} have driven substantial progress in multimodal understanding and zero-shot classification~\cite{li2025surveystateartlarge,CAO2025100069}. These models demonstrate impressive generalization to novel visual categories without task-specific training. However, their performance degrades on fine-grained tasks involving many semantically similar classes~\cite{zhang2024visuallygroundedlanguagemodelsbad,kim-ji-2024-finer, wei2024enhancingfinegrainedimageclassifications}. Even with advances in prompt learning~\cite{jha2024erapromptlearningvisionlanguage} and model adaptation~\cite{Cho_2023_ICCV, DBLP:journals/corr/abs-2411-16407}, challenges persist under dense class distributions, domain shifts, and nuanced language constructs like negation or ambiguity~\cite{alhamoud2025visionlanguagemodelsunderstandnegation, anis2025limitationsvisionlanguagemodelsunderstanding, gou2024visionlanguagemodelsimage}. These limitations motivate structured strategies to enhance model robustness and interpretability.

\textbf{Multi-label and Hierarchical Classification.}\\
A promising direction for improving visual classification involves structuring label spaces hierarchically to better guide visual decision-making and support the understanding of Compositional Language Concepts (CLC) in images. Tree-based and taxonomy-aware approaches break down complex objects into their attributes, states, and relationships with other objects in the scene, enhancing both prediction accuracy and interpretability in complex visual settings~\cite{yellinek20253vlusingtreesimprove, ding2025treeattributespromptlearning}. Structured label hierarchies allow models to reason step by step across different levels of abstraction~\cite{zhang2024improvevisionlanguagemodel}. Despite this potential, hierarchical method remain relatively underexplored in the context of image classification. Complementary strategies enrich label semantics by incorporating natural language descriptions generated by large language models (LLMs), which have been shown to improve zero-shot generalization in fine-grained classification tasks~\cite{saha2024improvedzeroshotclassificationadapting}. Nevertheless, VLM continue to face challenges in distinguishing subtly different or compositional classes, particularly in real-world scenarios involving occlusion, ambiguity, or overlapping class boundaries~\cite{xu2024benchmarkingzeroshotrecognitionvisionlanguage}.

\textbf{Structured Reasoning and Prompt Engineering}\\
To address these shortcomings, recent work has explored integrating structured reasoning and prompt engineering into VLM pipelines. Hierarchical or tree-structured reasoning provides a framework for decomposing classification into interpretable decision paths~\cite{besta2025demystifyingchainstreesgraphs, yellinek20253vlusingtreesimprove, wang2024videotree}. Prompt-based techniques, such as using category-specific templates or multi-step inference chains, help align model predictions with human-understandable logic~\cite{he-etal-2024-advancing}. In particular, prompt tuning enables VLMs to better navigate complex visual scenes, improving resilience to occlusion, perspective shifts, and context ambiguity~\cite{yellinek20253vlusingtreesimprove}. Recent studies further demonstrate that combining image-text reasoning with structured label spaces (e.g., via intermediate textual descriptions) can improve the accuracy over direct label prediction pipelines. These approaches support the development of debuggable, interpretable systems that maintain hierarchical consistency and allow for targeted analysis of model errors.

\section{Problem Statement}

The deployment of VLMs in real-world applications has led to increased interest in structured reasoning approaches that theoretically promise both improved interpretability and potentially enhanced performance. However, the practical effectiveness of these approaches compared to standard zero-shot classification methods remains an open empirical question.

\textbf{Challenge 1: Tree-based vs. Zero-shot Classification.} While VLMs have demonstrated impressive zero-shot capabilities, there is ongoing debate about whether incorporating explicit structural guidance through hierarchical reasoning (such as decision trees) provides additional benefits. Tree-based approaches offer theoretical advantages including interpretability, systematic error analysis, and the ability to decompose complex decisions into simpler     choices. However, they also introduce potential complications including error propagation through hierarchical structures, increased inference complexity, and sensitivity to tree design choices.

\textbf{Challenge 2: Performance-Interpretability Trade-offs.} The assumption that structured reasoning approaches necessarily improve performance while providing interpretability benefits has not been thoroughly validated empirically. It remains unclear whether the additional complexity introduced by hierarchical structures compensates for any potential accuracy gains, particularly given the strong performance of direct zero-shot methods.

\textbf{Challenge 3: Validation of Structured Knowledge.} Even when VLMs demonstrate understanding of the hierarchical knowledge embedded in decision trees, it is unclear whether this understanding translates into improved classification performance. The relationship between knowledge verification accuracy and actual classification effectiveness requires empirical investigation.\\

\textbf{Research Questions.} Our investigation centers on the following empirical questions:
\begin{enumerate}
\item \textbf{Performance Comparison:} How does tree-based hierarchical reasoning compare to standard zero-shot prompting in terms of classification accuracy across fine-grained and coarse-grained datasets?
\item \textbf{Knowledge Translation:} Is VLMs accurate in reasoning and answering visual questions about the datasets' classes? If yes, Does this reasoning or knowledge translate to improved classification performance in hierarchical reasoning systems?
\item \textbf{Error Analysis:} What are the primary failure modes of tree-based reasoning systems compared to direct zero-shot approaches?
\item \textbf{Robustness Assessment:} How do tree-based and zero-shot approaches compare in terms of consistency across different types of classification tasks?
\end{enumerate}

\section{Methodology}

\begin{figure*}[!t]
  \centering
  \includegraphics[trim=0pt 30pt 0pt 30pt, clip, width=0.75\linewidth]{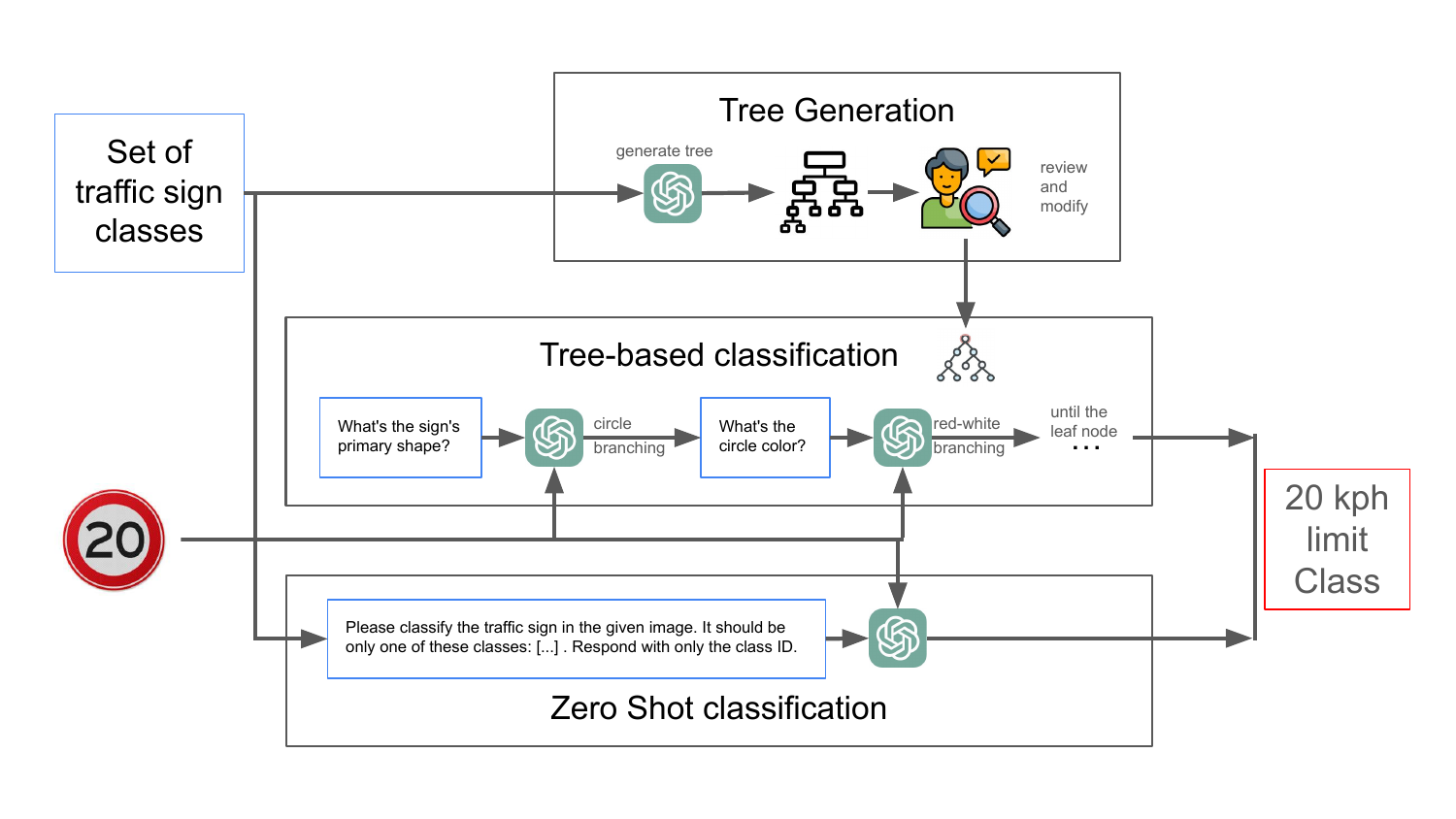}
  \caption{Experimental framework setup illustrating both zero-shot classification and hierarchical tree-based reasoning. It also shows the construction and use of decision trees for step-by-step visual classification.}
  \label{fig:visual-abstract}
\end{figure*}

To evaluate the tree-based inference approach, we construct decision trees for image classification across two real-world tasks. We experiment incorporating the history of previously traversed questions and answers into the tree model's prompt. We further augment the model input with class descriptions and image visual description generated by an LLM to provide richer semantic grounding. To assess the quality and interpretability of the generated decision trees, we conduct a verification study examining the VLM’s understanding of tree structure and branching logic. Figure~\ref{fig:visual-abstract} presents an overview of our experimental framework, including zero-shot classification, hierarchical reasoning, and the construction and use of decision trees for step-by-step visual classification.




\subsection{Datasets}
We evaluate the approach on two benchmark datasets, each representing a real-world image classification task. These datasets were chosen to balance task complexity with practical relevance:

\begin{itemize}
\item \textbf{German Traffic Sign Recognition Benchmark (GTSRB)} \cite{Stallkamp2012}:
\begin{itemize}
\item 43 traffic sign classes
\item Over 50,000 images of varying sizes
\item Sequential captures (multiple frames per sign)
\end{itemize}
This dataset introduces a fine-grained image classification task, requiring models to distinguish between visually similar German traffic signs under real-world, noisy conditions. Its relevance to safety-critical domains, such as autonomous driving, makes it a valuable testbed for evaluating hierarchical reasoning. To reduce redundancy, we sample \textbf{901 representative images} by selecting one random frame per sequence from the training set, including all the sequences inside the dataset.

\item \textbf{CIFAR-10} \cite{krizhevsky2009learning}:
\begin{itemize}
    \item 10 object classes (airplane, automobile, bird, cat, deer, dog, frog, horse, ship, truck)
    \item \textbf{1,000 balanced samples} (100 per class)
\end{itemize}
This dataset presents a coarse-grained classification task involving general object categories commonly encountered in daily life. It is designed to assess visual generalization capabilities across diverse semantic classes and tests the model's ability to differentiate between broad, well-known object types.

\item \textbf{MNIST} \cite{lecun2010mnist}:
\begin{itemize}
\item 10 handwritten digit classes (0-9)
\item \textbf{1,000 balanced samples} (100 per class)
\end{itemize}
This dataset represents a foundational fine-grained classification task, requiring models to distinguish between visually similar handwritten digits. Its inclusion allows us to assess the hierarchical reasoning framework on a task with low visual complexity but clear, distinct classes. We sampled \textbf{1,000 representative images} (100 per class) for our evaluation.

\item \textbf{MSCOCO (Microsoft Common Objects in Context)} \cite{lin2014microsoft}:
\begin{itemize}
\item 80 object categories
\item \textbf{850 balanced samples} (max 12 per category)
\item Images filtered for single-category content
\end{itemize}
This dataset provides a large-scale, coarse-grained object recognition task. MSCOCO images often contain complex scenes with multiple objects. To adapt this dataset for our single-label classification framework and ensure consistency with the other datasets, we applied a filtering process. We retained only those images containing exactly one object category, which resulted in a working set of \textbf{80 categories} (out of 90). From this filtered set, we sampled a maximum of 12 images per category to create our \textbf{850-sample balanced test set}.
\end{itemize}

\subsection{Models}
We conduct the evaluation experiments on three state-of-the-art VLMs, chosen to represent a range of architectural paradigms, size and capability levels. All models were used via API calls:

\begin{itemize}
    \item \textbf{GPT-4o} \\
    \textit{Rationale}: As OpenAI's most advanced multimodal model, GPT-4o provides a strong baseline for cutting-edge zero-shot visual reasoning. Its architecture excels at contextual understanding, making it ideal for testing hierarchical (tree-based) versus flat prompting strategies. \\
    \textit{Role in Study}: Serves as our primary model for cross-dataset comparison due to its robust performance.

    \item \textbf{LLaMA-3.2 11B Vision Instruct} \\
    \textit{Rationale}: This open-weight model balances scale and efficiency. Its decoder-only design contrasts with GPT-4o's hybrid architecture, testing generalization across model families. \\
    \textit{Role in Study}: Isolates the impact of model family (decoder-only vs. hybrid) and the model size variation. 

    \item \textbf{Qwen-VL MAX} \\
    \textit{Rationale}: A leading multilingual VLM with strong visual grounding capabilities, particularly in Asian contexts. Its performance on traffic signs (often designed with Unicode-like symbols) tests cultural and domain biases. \\
    \textit{Role in Study}: Clarify dataset-specific biases (e.g., GTSRB’s Western Unicode-like symbols) from model-specific limitations by cross-referencing performance against GPT-4o.

\end{itemize}

\subsection{Implementation Details}
Our methodology started with the construction of a hierarchical multi-decision tree tailored to each dataset to guide the classification process via a sequence of interpretable questions. The tree is traversed using depth-first search, and each leaf node maps to one final class label (e.g., \texttt{``20 kph speed limit''} or \texttt{``airplane''}). The models’ performance was rigorously evaluated by comparison with zero-shot prompting baselines corresponding to each model, as illustrated in Figure~\ref{fig:visual-abstract}.

\subsection{Tree Structure} 
The initial version of the tree is automatically generated using an LLM (We used both Deepseek and GPT-4.1), guided by class metadata and linguistic descriptions. The model is prompted to produce the tree in text format, with constraints ensuring no repeated questions along a single path and exactly one class at each leaf node. A human annotator then reviews, traverses, and refines the generated tree to ensure semantic accuracy and consistency.

A partial subtree~\footnote{The complete tree structure is available in the provided project code.} for both datasets used in the experiments is illustrated below:
\begin{lstlisting}
GTSRB Tree:
(Max Depth: 16, Number of nodes: 65)
[L0] Q: What's the sign's primary shape?
  -> triangle:
[L1] Q: Does the triangle have an exclamation mark?
      -> yes: Exclamation mark warning ([18])
      -> no: 
        [L2] Q: Does it depict a left curve?
          -> yes: Left curve warning ([19])
          -> no: (Continue traversal...)

CIFAR-10 Tree 
(Max Depth = 5, Number of nodes = 19)
[L2] Q: Is the animal typically shown with prominent hooves?
-> Yes (hooves):
[L3] Q: Does the animal have antlers visible in the image?
-> Yes (antlers): [L4] Leaf Node: deer (ID: 4)
-> No (no antlers): [L4] Leaf Node: horse (ID: 7)
-> No (paws): (Continue traversal...)
\end{lstlisting} 
\vspace{5pt}

\subsection{Decision Tree Knowledge Verification}

Before experimenting with tree-based classification, we first assessed two key components: the visual understanding capabilities of the strongest baseline VLM (GPT-4o) and the semantic coherence of our decision tree structure. To do this, we experimented using the decision tree built for the GTSRB dataset . In this setup, the model is given the ground truth class label and prompted to sequentially answer each question along the corresponding path in the tree that leads to this ground truth class. This allows us to evaluate both the model’s grasp of class-specific visual and semantic attributes, and the overall quality and structure of the decision tree, specifically, whether its questions are meaningful and well-organized for hierarchical classification.

For example, for the class ``20 kph speed limit'', the model interaction proceeds as follows:

\begin{lstlisting}[language=Python]
"20 kph speed limit": [
    {"[L0] Q: What's the sign's primary shape?": "circle"},
    {"[L1] Q: What's the circle color?": "red-white"},
    {"[L2] Q: Does it contain numbers?": "yes"},
    {"[L3] Q: What is the number?": "20"}
]
\end{lstlisting}
\vspace{5pt}

The model is given the class name up front and retains access to the full question-answer history to simulate a coherent reasoning process. We measure per-class accuracy as the number of correctly answered questions divided by the total number of questions along the path.

\subsection{Prompting Strategies}
We tested two main prompting strategies for visual classification across both datasets and all selected three VLMs, as illustrated below:

\begin{itemize}
    \item \textbf{Baseline (Zero-shot Classification)}:\\
    This is a standard prompting strategy that directly asks the model to classify the image by selecting a single class from the full classes set. The model receives the image as input along with the following prompt:
    
    \begin{lstlisting}[language=Python]
"Please classify the [Dataset object] in the given image. It should be only one of these classes: {class_ids_and_names}. Respond with only the class ID."
    \end{lstlisting}
    
    We also extend this method by providing Deepseek-generated textual descriptions for each class and a generated image caption by the tested model within the prompt context. We refer to this variation as \textbf{Zero-shot with Description Classification}.
    
    \item \textbf{Tree-Based Classification}:\\
    This is the primary strategy evaluated in our study. It involves traversing a decision tree from the root to a leaf node, where the classification decision is made. At each node, the prompt includes the node-specific question and a set of possible branching answers. The model is expected to choose one answer based on the image, which determines the next node to visit, continuing this process until reaching the final classification.
    
    The prompt template used at each node is as follows:
    
    \begin{lstlisting}[language=Python]
"{tree.question} Choose one of these answers: {tree.answers}."
    \end{lstlisting}
    
    An example from the GTSRB dataset tree:
    \begin{itemize}
        \footnotesize
        \item \textbf{Question:} What's the sign's primary shape?
        \item \textbf{Answers:} \scriptsize{[``triangle'', ``circle'', ``diamond'', ``inverted-triangle'', ``octagon'']}
    \end{itemize}
    Furthermore, we extended the tree-based strategy by introducing two additional variations:

    \begin{itemize}
        \item \textbf{Tree-Based Classification with History}:\\
        In this variation, we append the sequence of prior decisions made by the model to each prompt. This provides additional context about the path taken so far.
    
        \item \textbf{Tree-Based Classification with Descriptions}:\\
        Similar to the Zero-shot with Description strategy, we include LLM-generated textual descriptions of the classes and the target image caption as part of the prompt context. This information is provided before the model selects a branching answer, offering semantic guidance during classification.
    \end{itemize}

\end{itemize}

\subsection{Answer Extraction}
Due to the inherently non-deterministic nature of VLM responses, we cannot fully control their output format. To reliably extract answers, we employ regular expression pattern matching, which captures the first occurrence of any predefined answer candidate from the model's response.

\section{Evaluation and Results}
A comprehensive evaluation of the decision tree-based methodology is conducted and benchmarked against a zero-shot image classification baseline. Further details are provided in the following sections.

\subsection{Decision Tree Knowledge Verification}
We tested the GPT-4o model using the decision tree developed for the GTSRB dataset. The model achieved \textbf{100\% accuracy} on all decision tree questions for \textbf{39 out of 43} traffic sign classes, with an overall average accuracy of \textbf{98.20\%}. Crucially, the VLMs demonstrated comprehensive knowledge, correctly answering every tree branch question for each specific class. This validates that the curated decision tree's structure and question order are optimally suited for the task.

The four classes with slightly lower performance---\textit{``End of restriction''}, \textit{``Right-of-way at intersection''}, \textit{``Double curve warning''}, and \textit{``Ice/snow warning''}---exhibit high visual or semantic similarity to other signs, leading to occasional confusion. These results confirm the model's strong reasoning capabilities and the decision tree's alignment with the classification challenge. The minimal errors likely reflect ambiguities in visual design or contextual interpretation of certain signs rather than deficiencies in the tree itself.

\subsection{Zero-shot Classification}

\begin{figure}[ht]
    \centering
    \includegraphics[width=\linewidth]{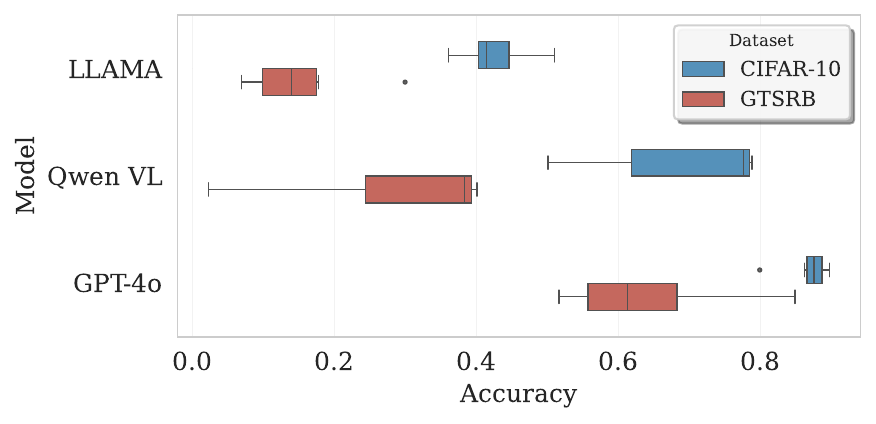}
    \caption{Distribution of zero-shot classification accuracy over 10 prompt variations, evaluated across GTSRB \& CIFAR-10.}
    \label{fig:prompt_sensitivity}
\end{figure}

We began our evaluation with a baseline zero-shot classification experiment. Given the non-deterministic behavior of VLMs and their known sensitivity to prompt phrasing~\cite{fatemi2023talklikegraphencoding}, we performed the classification using \textbf{10 diverse prompts} to assess robustness. The model was run with a \textbf{temperature of 0.7} to allow variability in responses. By incorporating prompt variation, we aimed to better understand the model's generalization and reduce performance artifacts caused by prompt-specific biases. Figure~\ref{fig:prompt_sensitivity} illustrates the distribution of zero-shot classification accuracy across different prompts.

\begin{figure*}[ht]
    \centering
    \includegraphics[width=\linewidth]{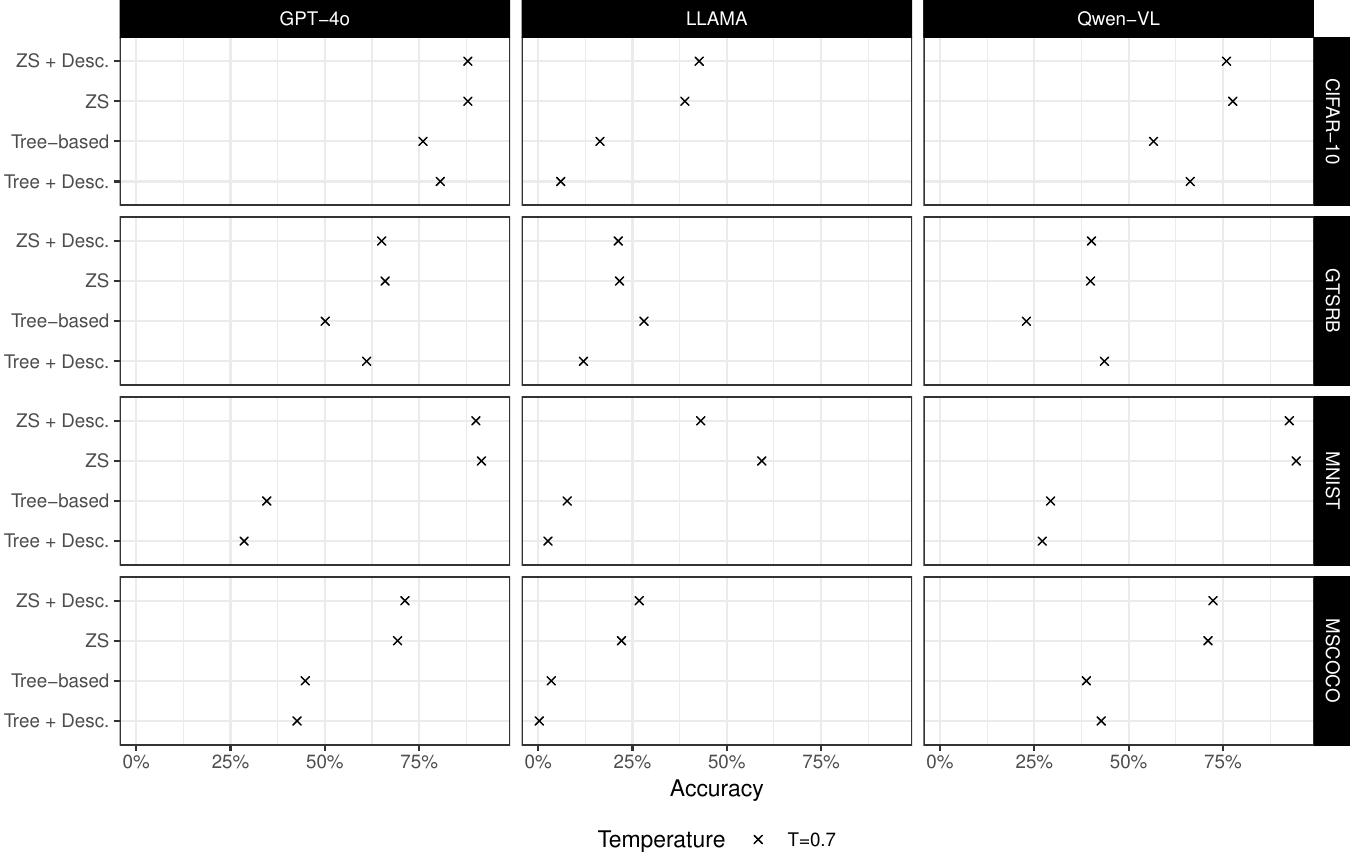}
    \caption{Mean accuracy (\%) comparison across prompting strategies, models and datasets.}
    \label{fig:results}
\end{figure*}

We then selected a \textbf{representative prompt} with average performance (based on mean accuracy across experiments) and conducted three runs for each of two temperature settings: \textbf{0.7 and 0}. The results were consistent across experiments, showing little to no variance. Figure~\ref{fig:results} presents the mean zero-shot accuracy across all classes.

Additionally, we ran a variant of the experiment that included \textbf{class descriptions} as part of the prompt. While this slightly improved performance for the LLaMA model on CIFAR-10, it did not benefit other experiments. For the GTSRB dataset, incorporating textual descriptions for the large number of classes increased context length \textbf{without meaningful accuracy gains}, suggesting that excessive prompt context may hinder performance in fine-grained classification tasks.

\subsection{Tree-Based Classification}


We evaluated the selected models on the generated decision trees using two temperature settings: \textbf{0.7 and 0}. For each setting, we recorded the \textbf{mean accuracy} across all classes. Figure~\ref{fig:results} presents the results for the tree-based classification approach.

\begin{itemize}
    \item \textbf{GPT-4o} consistently achieved the highest performance across both datasets and temperature settings, reaching \textbf{52.05\% accuracy} on GTSRB and \textbf{75.40\%} on CIFAR-10 (temperature 0). These results highlight the increased complexity of the decision tree in fine-grained classification tasks like GTSRB.
    \item In contrast, the \textbf{LLaMA model} exhibited relatively poor and unstable performance, particularly with varying temperature settings, suggesting lower reliability in following structured reasoning prompts.
\end{itemize}

Furthermore, Figure~\ref{fig:results} illustrates the impact of \textbf{two prompting variations} in the tree-based classification setting:

\begin{enumerate}
    \item \textbf{Including full decision path history} negatively affected performance, likely due to context overload with non-essential information.
    \item \textbf{Adding class descriptions} generally improved performance across datasets and models---except for LLaMA, where it caused a performance drop.
\end{enumerate}

These results suggest that \textbf{descriptive context enhances understanding in larger VLMs} capable of effectively leveraging long-context inputs.

\subsection{Accuracy Comparison}    


Contrary to our expectations, the experimental results indicate that incorporating tree-based reasoning structures does not improve performance for any of the evaluated VLMs on any of our fine-grained (GTSRB, MNIST) and coarse-grained (CIFAR-10, MSCOCO) classification tasks when compared to the zero-shot baseline. The results, illustrated in Figure~\ref{fig:results}, are centered on a temperature of 0.7. To ensure these findings are not artifacts of stochasticity, we also conducted all key experiments at a temperature of 0. These deterministic runs confirmed a consistent performance trend, validating that the superiority of zero-shot prompting is a robust finding and not dependent on temperature settings.

For the fine-grained GTSRB dataset (T=0.7), we observe a substantial performance gap. The best-performing model, GPT-4o, achieved 65.97\% accuracy in the zero-shot setting compared to just 50.10\% with tree-based reasoning. Similarly, Qwen-VL (T=0.7) showed a noticeable zero-shot superiority, with 39.88\% accuracy versus 22.89\% in the tree-based setup. These consistent gaps suggest that the models may not effectively leverage structured reasoning and instead rely more on dataset-specific biases than on decomposing and interpreting traffic signs' visual attributes.

This performance degradation is even more pronounced on the MNIST dataset (T=0.7). For this fine-grained task, the gap widens dramatically: Qwen-VL achieves 94.40\% accuracy in the zero-shot setting but plummets to 29.30\% with tree-based reasoning. Similarly, GPT-4o drops from 91.50\% (zero-shot) to just 34.60\% (tree-based), reinforcing the tree-based method's ineffectiveness.

The same trend persists in the coarse-grained tasks (T=0.7). For CIFAR-10, GPT-4o's accuracy dropped from 87.90\% (zero-shot) to 76.00\% (tree-based). This pattern is confirmed by the large-scale coarse-grained MSCOCO dataset, where Qwen-VL’s accuracy falls from 71.03\% (zero-shot) to 38.79\% (tree-based), and GPT-4o’s drops from 69.27\% to 44.81\%.

Furthermore, attempts to augment the tree-based method, such as including the full decision path history in the prompt (Chat Mode), failed to improve results. This variation often degraded performance further, likely due to increasing the context length with non-essential information without providing meaningful guidance. Notably, per-class analysis shows the zero-shot approach outperformed tree-based classification across all CIFAR-10 classes, and on GTSRB, the tree-based method was only superior for 11 of the 43 classes, on MNIST, the zero-shot across all the experiments outperformed in nearly all the classes(93.34\% of the experiments), and MSCOCO, the zero-shot across all the experiments outperformed in nearly all the classes(91.67\% of the experiments).

These findings suggest that the additional complexity of hierarchical reasoning structures does not enhance VLM performance in vision-related classification tasks. Instead, the simpler zero-shot baseline proves more effective for both fine-grained (GTSRB, MNIST) and coarse-grained (CIFAR-10, MSCOCO) settings. We explore potential explanations for these results in the Discussion section.

\section{Discussion}
\label{sec:discussion}

Our experimental results demonstrate consistent performance degradation when employing tree-based reasoning compared to the zero-shot baseline, revealing several critical contributing factors. The tree-based approach exhibits significant sensitivity to errors in upper-level decision nodes, where top-level fragility proves particularly problematic. 

In the GTSRB dataset, preliminary questions such as ``Does it contain a number?'' disproportionately affect downstream classification when answered incorrectly, while alternative question orderings yield accuracy variations that indicate strong path dependence in the reasoning process. Furthermore, structural deficiencies in tree design introduce substantial limitations, including question hierarchy flaws where parent nodes improperly subsume children (e.g., Layer 10: ``Pedestrian?'' versus Layer 11: ``Pedestrian with child?''), resulting in systematic misclassification of all ``pedestrian with child'' images as generic pedestrians.

The reasoning process also struggles with contextual integrity, as including conversation history degrades performance by introducing irrelevant information that dilutes classification-relevant features. This finding aligns with \cite{liu2023lost}'s demonstration that performance drops when input exceeds optimal context length or when critical information is suboptimally positioned.

The least accurate layers predominantly involve higher-level reasoning, where questions pertain to the semantic interpretation of traffic signs, such as their class or function, rather than low-level visual features like shapes or colors. This pattern suggests that the models perform reasonably well on basic perceptual tasks but demonstrate significant weaknesses when reasoning about abstract or contextual aspects, highlighting a fundamental limitation in current vision-language models regarding hierarchical reasoning capabilities.

Additionally, we observe that complex semantic queries (e.g., "What action is prohibited?") consistently underperform simple visual questions (e.g., "Triangular shape?"), further supporting the notion that current models excel at atomic visual recognition but struggle with semantic interpretation tasks requiring deep traversal of hierarchical structures. These findings suggest several architectural improvements for future implementations: breadth-first prioritization where shallow, wide trees outperform deep ones, and visual primitives first approaches where atomic visual queries concerning color and shape should precede semantic interpretation to maximize classification accuracy and minimize error propagation through the reasoning hierarchy.
\section{Conclusion}

Our study demonstrates that tree-based reasoning structures do not improve the performance of VLMs on vision-related classification tasks, as evidenced by experiments across four diverse datasets: GTSRB, MNIST, CIFAR-10, and MSCOCO. Instead, we observe a consistent and significant degradation in accuracy compared to standard zero-shot prompting across all evaluated models. Zero-shot prompting decisively outperforms tree-based reasoning in both fine-grained (GTSRB, MNIST) and coarse-grained (CIFAR-10, MSCOCO) settings, achieving higher mean accuracy and correctly classifying samples that the hierarchical approach misclassifies. These findings, now validated across multiple benchmarks, suggest that the added complexity of tree-based reasoning does not enhance VLM performance in such tasks and, in fact, introduces critical inefficiencies.

Furthermore, our experiments reveal that the tree-based methodology is highly sensitive to structural and question-formatting adjustments, indicating that minor changes in the reasoning path or question phrasing can significantly impact performance. Critically, the knowledge verification confirms VLMs possess a strong understanding of elementary visual attributes (achieving 98.20\% average accuracy across GTSRB classes), but their inability to reliably use this decomposed knowledge within the reasoning tree to arrive at the correct final classification provides a key explanation for the observed performance degradation. This failure to integrate elementary visual understanding through structured decomposition suggests fundamental limitations in the compositional reasoning capabilities of current VLMs for such classification tasks, rather than a simple inefficiency in the method.

\subsection{Future Work}

    
    


To enhance the effectiveness of tree-based reasoning in vision-language tasks, several promising directions warrant further exploration. One approach involves the investigation of hybrid reasoning strategies that integrate zero-shot predictions into the tree-based reasoning pipeline. For example, incorporating the zero-shot label as an additional feature in the decision tree could help reduce misclassifications and capitalize on the complementary strengths of both methods.

In addition, future work could focus on developing automated or semi-automated techniques for generating optimal decision trees. Methods such as reinforcement learning or iterative refinement may prove effective in constructing trees that are both accurate and robust, potentially reducing sensitivity to structural variations and enhancing generalizability. It is also important to explore whether these improvements hold across different multimodal tasks and datasets. Furthermore, alternative reasoning frameworks that strike a balance between interpretability and performance merit investigation. Addressing these challenges will contribute to a deeper understanding of when and how structured reasoning can amplify the capabilities of large language models in vision-language applications.
{
    \small
    \bibliographystyle{ieeenat_fullname}
    \bibliography{main}
}

\end{document}